\documentclass[letterpaper, 10 pt, conference]{ieeeconf}  

\IEEEoverridecommandlockouts                              

\usepackage{epsfig}
\usepackage{graphicx}
\usepackage{amsmath}
\usepackage{amssymb}
\usepackage{color, soul}

\usepackage[labelformat=simple]{subcaption}

\usepackage{multirow}
\usepackage[breaklinks=true,bookmarks=false]{hyperref}
\usepackage{cleveref}

\usepackage{booktabs}
\usepackage{comment}
\usepackage{makecell}

\hypersetup{
    colorlinks=true,
    linkcolor=blue,
    filecolor=magenta,      
    urlcolor=magenta,
}

\newcommand{\xhdr}[1]{\vspace{5pt} \noindent {\textbf{#1} }}
\newcommand{\etal}{\textit{et al}.}

\newcommand{\norm}[1]{\left\lVert#1\right\rVert}

\title{\LARGE \bf
BiTraP: Bi-directional Pedestrian Trajectory Prediction with Multi-modal Goal Estimation
}

\author{Yu Yao$^{1}$, Ella Atkins$^{2}$, Matthew Johnson-Roberson$^{3}$, Ram Vasudevan$^{4}$, and Xiaoxiao Du$^{3}$
\thanks{This work was supported by a grant from Ford Motor Company via the Ford-UM Alliance and the Federal Highway Administration. }
\thanks{$^{1}$Y. Yao is with the Robotics Institute, University of Michigan, Ann Arbor, MI 48109 USA
        {\tt\footnotesize brianyao@umich.edu}}%
\thanks{$^{2}$E. Atkins is with the Aerospace Engineering Department, University of Michigan, Ann Arbor, MI 48109 USA
        {\tt\footnotesize ematkins@umich.edu}}
\thanks{$^{3}$M. Johnson-Roberson and X. Du are with Department of Naval Architecture and Marine Engineering, University of Michigan, Ann Arbor, MI 48109 USA
        {\tt\footnotesize  mattjr@umich.edu;
        xiaodu@umich.edu}}%
\thanks{$^{4} $R. Vasudevan is with the Department of Mechanical Engineering, University of Michigan, Ann Arbor, MI 48109 USA
        {\tt\footnotesize ramv@umich.edu}}%
}

\begin{document}

\maketitle
\thispagestyle{empty}
\pagestyle{empty}

\begin{abstract}

Pedestrian trajectory prediction is an essential task in robotic applications such as autonomous driving and robot navigation. 
State-of-the-art trajectory predictors use a conditional variational autoencoder (CVAE) with recurrent neural networks (RNNs) to encode observed trajectories and decode multi-modal future trajectories. This process can suffer from accumulated errors over long prediction horizons ($\geq2$ seconds).
This paper presents \textit{BiTraP}, a goal-conditioned bi-directional multi-modal trajectory prediction method based on the CVAE. BiTraP estimates the goal (end-point) of trajectories and 
introduces a novel bi-directional decoder to improve longer-term trajectory prediction accuracy.
Extensive experiments show that BiTraP generalizes to both first-person view (FPV) and bird's-eye view (BEV) scenarios and outperforms state-of-the-art results by $\sim10-50\%$.
We also show that different choices of non-parametric versus parametric target models in the CVAE directly influence the predicted multi-modal trajectory distributions.
These results provide guidance on trajectory predictor design for robotic applications such as collision avoidance and navigation systems.\textit{ Our code is available at: \url{https://github.com/umautobots/bidireaction-trajectory-prediction}}

\end{abstract}

\section{Introduction}
Understanding and predicting pedestrian movement behaviors is crucial for autonomous systems to safely navigate interactive environments. By correctly forecasting pedestrian trajectories, a robot can plan safe and socially-aware paths in traffic~\cite{alahi2016social,liang2019peeking,sivaraman2014dynamic,li2019game} and produce alarms about anomalous motions (e.g., crashes or near collisions)~\cite{morais2019learning,yao2019unsupervised,yao2020dota,yao2018smart,yao2020smart}. Early work often assumed a deterministic future, where only one trajectory is predicted for each person given past observations \cite{kalman1960new, helbing1995social, williams2006gaussian}. However, pedestrians move with a high degree of stochasticity so multiple plausible and distinct future behaviors can exist \cite{gupta2018social,fragkiadaki2015recurrent}. Recent studies \cite{jacobs2017real,lee2017desire,anderson2019stochastic,chai2019multipath,ivanovic2019trajectron, salzmann2020trajectron++} have shown predicting a distribution of multiple potential future trajectories (i.e., multi-modal prediction) rather than a single best trajectory can more accurately model future motions of pedestrians.

Recurrent neural networks (RNNs), notably long short-term memory networks (LSTMs) and gated recurrent units (GRUs), have demonstrated success in trajectory prediction~\cite{liang2019peeking, du2019bio,yao2019egocentric,rasouli2019pie}. 
However, existing models recurrently predict future trajectories based on previous output thus their performance tends to deteriorate rapidly over time ($>$ 560 ms)~\cite{fragkiadaki2015recurrent, butepage2018anticipating}. 
We propose to address this problem with a novel goal-conditioned bi-directional trajectory predictor, named \textit{BiTraP}. BiTraP first estimates future goals (end-points of the future trajectories) of pedestrians and then predicts trajectories by combining forward passing from current position and backward passing from estimated goals.
We believe that predicting goals can improve long-term trajectory predictions, as pedestrians in real world often have desired goals and plan paths to reach these goals~\cite{mangalam2020endpoint}. Compared to existing goal-conditioned methods~\cite{mangalam2020endpoint,rehder2015goal,rhinehart2019precog} where goals were used as an input to a forward decoder, BiTraP takes goals as the starting position of a backward decoder and predicts future trajectories from two directions, thus mitigating the accumulated error over longer prediction horizons.

Recently, generative models such as the generative adversarial network (GAN) \cite{gupta2018social} and conditional variational autoencoder (CVAE)~\cite{sohn2015learning, lee2017desire}, were developed to predict multi-modal distributions of future trajectories. Our BiTraP model predicts multi-modal trajectories based on CVAE which learns target future trajectory distributions conditioned on the observed past trajectories through a stochastic latent variable. The two most common forms of the latent variable follow either a Gaussian distribution or a categorical distribution, resulting in either a non-parametric target distribution \cite{lee2017desire,mangalam2020endpoint} or a parametric target distribution model
such as a Gaussian Mixture Model (GMM)~\cite{ivanovic2019trajectron,salzmann2020trajectron++}. 
There has been limited research on how latent variable distributions impact predicted multi-modal trajectories. To fill this gap, we conducted extensive comparison studies using two variations of our BiTraP method: a non-parametric model using Gaussian latent variables (BiTraP-NP) and a GMM model using categorical latent variables (BiTraP-GMM). We implemented two types of loss functions, best-of-many (BoM) L2 loss~\cite{bhattacharyya2018accurate} and negative log-likelihood (NLL) loss~\cite{salzmann2020trajectron++} to evaluate different predicted trajectory behaviors (e.g., spread and diversity). We show that latent variable distribution choices are closely related to the diversity of predicted distributions, which provides guidance for selecting trajectory predictors for robot navigation and collision avoidance systems.

The contributions of this work are summarized as follows. First, we developed a novel bi-directional trajectory predictor, \textit{BiTraP}, based on multi-modal goal estimation and show it offers significant improvements on trajectory prediction performance especially for longer ($\geq2$ seconds) prediction horizons.  
Second, we studied parametric versus non-parametric target modeling methods 
by presenting two variations of our model, BiTraP-NP and BiTraP-GMM, and compare their influence on the diversity of predicted distribution.
Extensive experiments with both first person and bird's eye view datasets 
show the effectiveness of BiTraP models in different domains. 

\section{Related Work}

Our BiTraP model consists of two parts: a multi-modal goal estimator and a goal-conditioned bi-directional trajectory predictor. This section describes related work in multi-modal trajectory prediction and goal-conditioned prediction.

\xhdr{CVAE-based Approaches for Multi-modal Trajectory Prediction.} 
Probabilistic approaches, particularly conditional variational autoencoder (CVAE) based models, have been developed for multi-modal trajectory prediction. Different from GANs~\cite{gupta2018social,kosaraju2019social}, CVAEs can explicitly learn the form of a target distribution conditioned 
on past observations by learning the latent distribution from which it samples. 
Some CVAE methods assume the target trajectory follows a non-parametric (NP) 
distribution and produces multi-modal predictions by sampling from a Gaussian 
latent space. Lee~\etal~\cite{lee2017desire} first used CVAE for multi-modal 
trajectory prediction by incorporating Gaussian latent space sampling 
to an long short-term memory encoder-decoder (LSTM-ED) model. CVAE with LSTM components has since been used in many applications~\cite{deo2018multi,ivanovic2018generative,choi2019drogon}. Other CVAE-based methods assume parametric trajectory distributions. Ivanovic~\etal\cite{ivanovic2019trajectron} assumed the target
trajectory follows a Gaussian Mixture Model (GMM) and designed a Trajectron 
network to predict GMM parameters using a spatio-temporal graph.
Trajectron++~\cite{salzmann2020trajectron++} extended Trajectron to account for
dynamics and heterogeneous input data. Our work extends existing CVAE models to include goal estimation and shows improved multi-modal prediction results.  Our work also provides novel insights in comparisons between CVAE target distributions (NP and GMM).

\xhdr{Trajectory Conditioned on Goals. }
Incorporating goals has been shown to improve trajectory prediction. Rehder~\etal~\cite{rehder2015goal} proposed a particle-filter based method to 
estimate goal distribution as a prior for trajectory prediction.
We drew inspiration from~\cite{rehder2018pedestrian}, which computed  forward and backward rewards based on current position and goal; the path is planned using Inverse Reinforcement Learning (IRL). 
Our work is distinct due to its bi-directional temporal propagation 
and integration combined with a CVAE to achieve multi-modal prediction.
Rhinehart~\etal~\cite{rhinehart2019precog} estimated multi-modal semantic action as goals and planned conditioned trajectories using imitative models.
Deo~\etal~\cite{deo2020trajectory} used IRL to estimate goal states and fused results with past trajectory encodings to generate predictions.
Most recently, Mangalam~\etal~\cite{mangalam2020endpoint} designed a PECNet which showed state-of-the-art results on BEV trajectory prediction datasets. However, PECNet only concatenated past trajectory encodings and end-point encodings, which we believe did not fully take advantage of goal information. We have designed a bi-directional trajectory decoder in which current trajectory information is passed forward to the end-points (goals) and goals are recurrently propagated back to the current position. Experiment results show that our goal estimation can help generate more accurate trajectories. 


\section{BiTraP: Bi-directional Trajectory Prediction with Goal Estimation}

Our BiTraP model performs goal-conditioned multi-modal bi-directional trajectory prediction in either first-person view (FPV) or bird's eye view (BEV).
Let $\mathbf{X}_t=[X_{t-\tau+1},X_{t-\tau+2},...,X_{t}]$ denote observed past trajectory at time $t$, where $X_t$ is bounding box location and size $(x,y,w,h)$ in pixels for FPV~\cite{yao2019egocentric,rasouli2019pie} and $(x,y)$ position in meters for BEV~\cite{salzmann2020trajectron++}. Given $\mathbf{X}_t$, we first estimate goal $G_t$ of the person then predict future trajectory 
$\mathbf{Y}_t=[Y_{t+1},Y_{t+2},...,Y_{t+\delta}]$, where $\tau$ and 
$\delta$ are observation and prediction horizons, respectively. Define goal $G_t=Y_{t+\delta}$ as the future trajectory endpoint, which is given in training and unknown in testing.
We adopt a CVAE model to realize multi-modal goal and trajectory prediction. BiTraP contains four sub-modules: {conditional prior network} $p_\theta(Z|\mathbf{X}_t)$ to model latent variable $Z$ from observations,  {recognition network} $q_\phi(Z|\mathbf{X}_t, \mathbf{Y}_{t})$ to capture dependencies between $Z$ and $\mathbf{Y}_{t}$, {goal generation network}  $p_\omega(G_{t}|\mathbf{X}_t, Z)$, and {trajectory generation network} $p_\psi(\mathbf{Y}_{t}|\mathbf{X}_t, G_{t}, Z)$ where $\phi$, $\theta$, $\omega$ and $\psi$ represent network parameters. Either parametric or non-parametric models can be used to design networks
$p_{\psi}$ and $p_{\omega}$ for CVAE. Non-parametric models do not assume the distribution
format of target $\mathbf{Y}_t$ but learn it implicitly by learning
the distribution of $Z$. Parametric models assume a known distribution
format for $\mathbf{Y}_t$ and predict distribution parameters. 
We design non-parametric and parametric models in
Sections~\ref{sec:bitrap-np} and~\ref{sec:bitrap-gmm}, 
and explain different loss functions to train these models in Sections~\ref{sec:bitrap-np-loss} and~\ref{sec:bitrap-gmm-loss}.

\subsection{BiTraP with Non-parametric (NP) Distribution}\label{sec:bitrap-np}

\begin{figure*}[h!]
    \centering
    \includegraphics[width=0.95\textwidth]{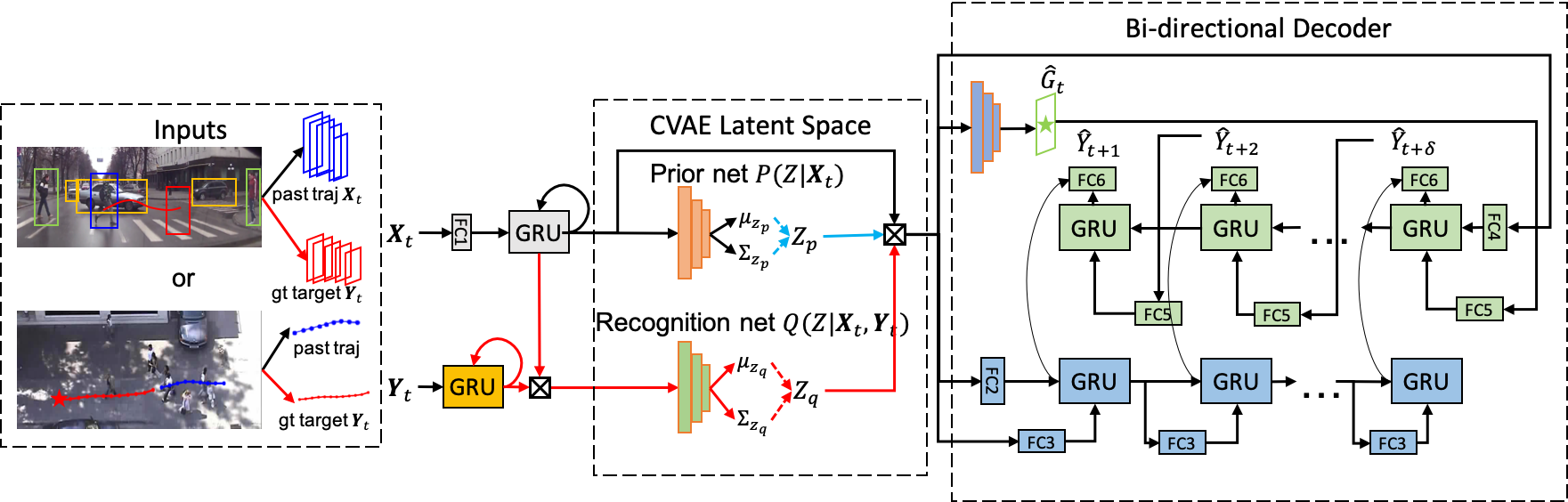}
    \caption{Overview of our BiTraP-NP network. Red, blue and black arrows show processes that appear in training only, inference only, and both training and inference, respectively.}
    \label{fig:bitrap_np}
\end{figure*}

BiTraP-NP is built on a standard recurrent neural network encoder-decoder (RNN-ED) based CVAE trajectory predictor as in~\cite{lee2017desire,mangalam2020endpoint,bhattacharyya2018accurate,ivanovic2018generative}, except it predicts goal first and then predict trajectories leveraging goals. Following previous work, we
assume Gaussian latent variable $Z\sim\mathcal{N}(\mu_{Z}, \sigma_{Z})$ and a non-parametric target distribution format.
Fig.~\ref{fig:bitrap_np} shows the network architecture of BiTraP-NP. 

\xhdr{Encoder and goal estimation.} 
First, observed trajectory $\mathbf{X}_t$ is processed by a gated-recurrent unit (GRU) encoder network to obtain encoded feature vector $h_t$.
In training, ground truth target $\mathbf{Y}_t$ is 
encoded by another GRU yielding $h_{Y_t}$. Recognition network
$q_{\phi}(Z|\textbf{X}_t,\textbf{Y}_t)$ takes $h_t$ and $h_{Y_t}$ to 
predict distribution mean $\mu_{Z_q}$ and covariance $\Sigma_{Z_q}$ 
which capture dependencies between observation and ground 
truth target. 
Prior network $p_{\theta}(Z|\textbf{X}_t)$ assumes no knowledge about 
target and predicts $\mu_{Z_p}$ and $\Sigma_{Z_p}$ 
using $h_t$ only. 
Kullback–Leibler divergence ($KLD$) loss between $\mathcal{N}(\mu_{Z_p}, 
\Sigma_{Z_p})$ and $\mathcal{N}(\mu_{Z_q},\Sigma_{Z_q})$ is optimized
so that dependency between $\mathbf{Y}_t$ and $\mathbf{X}_t$ is
implicitly learned by the prior network. Latent variable $Z$ is sampled from $\mathcal{N}(\mu_{Z_q},\Sigma_{Z_q})$ and concatenated with $h_t$ to predict multi-modal goals $\hat{G}_t$ with goal generation network $p_{\omega}(G_t|\mathbf{X}_t, Z)$. 
In testing, 
we directly draw multiple samples from $\mathcal{N}(\mu_{Z_p}, \Sigma_{Z_p})$
and concatenate $h_t$ to predict estimated goals $\hat{G}_t$. 
We use 3-layer multi-layer perceptrons (MLPs) for {prior}, {recognition} and {goal generation networks}.

\xhdr{Trajectory Decoder.} 
Predicted goals $\hat{G}_t$ are used as inputs to a bi-directional {trajectory generation network} $p_{\psi}(\mathbf{Y}_t|\mathbf{X}_t, \hat{G}_t, Z)$, the trajectory decoder, to predict multi-modal trajectories.
BiTraP's decoder contains forward and backward RNNs. The forward RNN is similar to a regular RNN decoder (Eq.~\eqref{eq:forward_rnn}) except its output is not transformed to trajectory space. The backward RNN is initialized from encoder hidden state $h_t$. It takes  estimated goal $\hat{Y}_{t+\delta}=\hat{G}_{t}$ as the initial input (Eq.~\eqref{eq:backward_rnn}) and propagates from time $t+\delta$ to $t+1$ so backward hidden state is updated from the goal to the current location. Forward and backward hidden states for the same time step are concatenated to predict the final trajectory way-point at that time (Eq.~\eqref{eq:bitrap_output}). These steps can be formulated as
\begin{align}
    & h_{t+1}^{f} = GRU_{f}(h_t^f, W_{f}^{i}h_t^f+b_{f}^i), \label{eq:forward_rnn}\\
    & h_{t+\delta-1}^{b} = GRU_{b}(h_{t+\delta}^b, W_{b}^{i}\hat{Y}_{t+\delta}+b_{b}^i), \label{eq:backward_rnn}\\
    &\hat{Y}_{t+\delta-1} = W_f^o h_{t+\delta-1}^f + W_b^o h_{t+\delta-1}^b + b^o, \label{eq:bitrap_output}
\end{align}
where, $f$, $b$, $i$ and $o$ indicate ``forward", ``backward", ``input" and ``output" respectively, and $h_t^f$ and $h_{t+\delta}^b$ are initialized by passing $h_t$ through two different fully-connected networks.



\subsection{BiTraP with GMM Distribution}\label{sec:bitrap-gmm}
Parametric models predict trajectory distribution parameters instead of trajectory coordinates. BiTraP-GMM is our parametric variation of BiTraP assuming a GMM for the trajectory goal and at each way-point~\cite{ivanovic2019trajectron,salzmann2020trajectron++}. Let $p(Y_{t+\delta})$ denote a $K$-component GMM at time step $t+\delta$. We assume $p(Y_{t+\delta})$ $= \sum_{i=1}^{K}\pi_i\mathcal{N}(Y_{t+\delta}|\mu_{t+\delta}^i, \Sigma_{t+\delta}^i)$, 
where each Gaussian component can be considered the distribution of one trajectory
modality. Mixture component weights $\pi_i$ sum to 
one, thus forming a categorical distribution. Each $\pi_i$ indicates 
the probability (confidence) that a person's motion belongs to that modality.
%
%
We design latent 
vector $Z$ as a categorical ($Cat$) variable $Z\sim Cat(K, \pi_{1:K})$ 
parameterized by GMM component weights $\pi_{1:K}$ rather than separately-computed parameters.
Similar to BiTraP-NP, we use three 3-layer MLPs for the {prior}, 
{recognition} and {goal generation networks}, and a bi-directional RNN decoder 
for the {trajectory generation network}. 
Instead of directly predicting trajectory coordinates,  generation networks of BiTraP-GMM estimate the $\mu_{t+\delta}^i$ and $\Sigma_{t+\delta}^i$ of the $i$th Gaussian components at time $t+\delta$. 
In training, we sample one $Z$ from each category to ensure all trajectory modalities are trained. 
In testing, we sample $Z$ from $Cat(K, \pi_{1:K})$ so it is more probable to sample from high-confidence trajectory modalities.

\subsection{Residual Prediction and BoM Loss for BiTraP-NP} \label{sec:bitrap-np-loss}
Instead of directly predicting future location~\cite{rasouli2019pie} or integrating from predicted future velocity~\cite{salzmann2020trajectron++}, BiTraP-NP predicts change with respect to the current location based on residuals $\hat{Y}_{t+\delta} = Y_{t+\delta} - X_t$. There are two advantages of residual prediction. First, it assures the model will predict the trajectory starting from the current location, providing smaller initial loss than predicting location from scratch. Second, the residual target can be less noisy than the velocity target due to the fact that trajectory annotation is not always accurate. 
Standard CVAE loss includes NLL loss of the predicted distribution which is not applicable to NP methods due to their unknown distribution format. L2 loss between predictions and targets can be used as a substitution~\cite{lee2017desire}. To further encourage diversity in multi-modal prediction, we use best-of-many (BoM) L2 loss as in~\cite{bhattacharyya2018accurate}. 
The final loss function for BiTraP-NP is a combination of the goal L2 loss, the trajectory L2 loss and the KL-divergence loss between prior and recognition networks, written as
\begin{align}
\begin{split}
    L_{NP} &= \min_{i\in N} \norm{G_t-X_t-\hat{G}_t^i} \\
    &+ \min_{i\in N} \sum_{\tau=t+1}^{t+\delta}\norm{Y_{\tau}-X_t-\hat{Y}_{\tau}^i} + KLD,
    \end{split}
    \label{eq:gaussian_loss}
\end{align}
where $\hat{G}_t$ and $\hat{Y}_\tau$ are the predicted goal and trajectory waypoints with respect to current position $X_t$.

\begin{figure*}[ht!]
    \centering
    \includegraphics[width=0.78\textwidth]{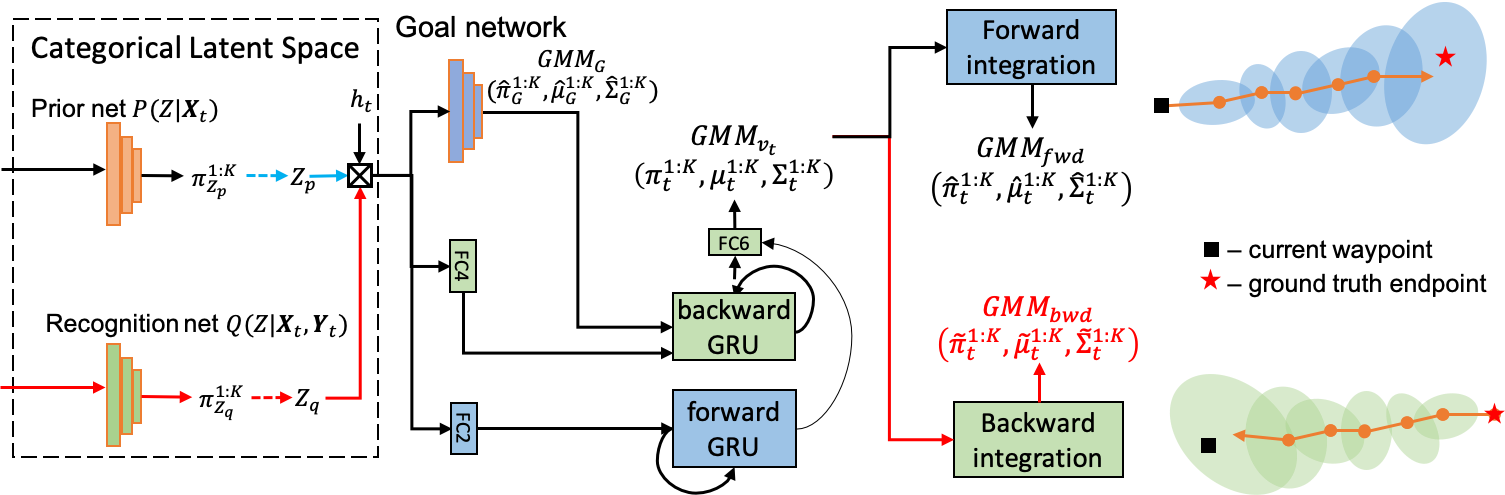}
    \caption{Latent space sampling and decoder modules of BiTraP-GMM. The ellipse shows one of $K$ GMM components at each timestep. The rest of the network is the same as BiTraP-NP in Fig.~\ref{fig:bitrap_np}. }
    \label{fig:Cat_GMM_diagram}
\end{figure*}

\subsection{Bi-directional NLL Loss for BiTraP-GMM} \label{sec:bitrap-gmm-loss}
Similar to \cite{salzmann2020trajectron++}, our BiTraP-GMM 
models the pedestrian velocity distribution as a GMM at each time step. 
The velocity GMM is then integrated forward to obtain the GMM distribution of trajectory waypoints $Y_{t+\delta}$ as shown 
by blue blocks in Fig.~\ref{fig:Cat_GMM_diagram}. We assume linear dynamics for pedestrian and use a single integrator as in Eq.~\eqref{eq:fwd_integrate}.
The loss function is then the summation of negative log-likelihood (NLL) 
of the ground truth future waypoints over the prediction horizon, formulated as 
\begin{align}
\begin{split}
   & GMM_{Y_{t+\delta}}(\hat{\pi}_{t+\delta}^{1:K}, \hat{\mu}_{t+\delta}^{1:K}, \hat{\Sigma}_{t+\delta}^{1:K}) \\
   &= X_t + \int_{t}^{t+\delta} GMM_{v_\tau}(\pi_{\tau}^{1:K}, \mu_{\tau}^{1:K}, \Sigma_{\tau}^{1:K})d\tau, \label{eq:fwd_integrate}
        \end{split}
\end{align}
\vspace{-5mm}
\begin{align}
\begin{split}
    NLL_{fwd} = \sum_{\tau=t}^{t+\delta}-\log{p(Y_{\tau}|\hat{\pi}_{\tau}^{1:K},\hat{\mu}_{\tau}^{1:K}, \hat{\Sigma}_{\tau}^{1:K}),}
    \label{eq:fwd_integrate_nll}
     \end{split}
\end{align}
where $\pi_{\tau}^{1:K},\, \mu_{\tau}^{1:K},\, \Sigma_{\tau}^{1:K}$ are velocity GMM parameters at time $\tau\in[t+1,t+\delta]$, and the $\hat{\cdot}$ symbol indicates location GMM parameters obtained from integration. $p(\cdot)$ is the $GMM$ probability density function. 
Such an $NLL$ emphasizes earlier waypoints along the prediction horizon because a waypoint at time $t+1$ is used in integration results over $t+2,\,t+3,...$, while these later waypoints are not used when computing $t+1$. This goes against our proposed idea which is to leverage a bi-directional temporal model. Therefore, we compute bi-directional NLL loss with reverse integration from the goal, formulated as
\begin{align}
\begin{split}
    &GMM'_{Y_{t}}(\tilde{\pi}_t^{1:K}, \tilde{\mu}_t^{1:K}, \tilde{\Sigma}_t^{1:K}) \\
    &= G_{t} - \int_{t+\delta}^{t} GMM_{v_\tau}(\pi_{\tau}^{1:K}, \mu_{\tau}^{1:K}, \Sigma_{\tau}^{1:K})d\tau, 
    \end{split}
\end{align}
\vspace{-5mm}
\begin{align}
\begin{split}
    NLL_{bwd} = \sum_{\tau=t+\delta}^{t}-\log{p'(Y_{\tau}|\tilde{\pi}_{\tau}^{1:K},\tilde{\mu}_{\tau}^{1:K}, \tilde{\Sigma}_{\tau}^{1:K}).}
    \end{split}
\end{align}
where $p(\cdot)'$ is the backward $GMM$ probability density function, the $\tilde{\cdot}$ symbol indicates backward location GMM parameters.  
The final loss function for BiTraP-GMM can be written as
\begin{align}
\begin{split}
    L_{GMM} &=  -\log{p_{_G}(G_{t}|\hat{\pi}_G^{1:K},\hat{\mu}_G^{1:K}, \hat{\Sigma}_G^{1:K})} \\
    &+ NLL_{fwd} + NLL_{bwd} + KLD, \label{eq:gmm_loss}
\end{split}
\end{align}
where the first term is $NLL$ loss of the goal estimation, $NLL_{fwd}$ and $NLL_{bwd}$ are computed from forward and backward integration, the $KLD$ term is the KL-divergence similar to Eq.~\eqref{eq:gaussian_loss}.


\section{Experiments and Results}\label{sec:experiments}
In this section, we empirically evaluate BiTraP-NP and BiTraP-GMM models on both 
first-person view (FPV) and bird's eye view (BEV) trajectory prediction datasets. We also provide a comparative study and discussion on the effects of model and loss selection.

\xhdr{Datasets.} Two
FPV datasets, Joint Attention for Autonomous Driving (JAAD) 
~\cite{kotseruba2016jaad} and Pedestrian Intention 
Estimation (PIE) ~\cite{rasouli2019pie}, and two benchmark BEV 
datasets, ETH~\cite{pellegrini2009eth} and UCY~\cite{leal2014ucy}, were used in our experiments. 
JAAD contains 2,800 pedestrian trajectories captured from dash cameras annotated at 30Hz. PIE contains 1,800 pedestrian 
trajectories also annotated at 30Hz, with longer trajectories and more comprehensive
annotations such as semantic intention, ego-motion and neighbor objects. 
ETH-UCY datasets contain five sub-datasets captured from 
down-facing surveillance cameras in four different scenes with 1,536  pedestrian trajectories annotated at 2.5Hz.

\xhdr{Implementation Details.} We used the standard training/testing splits of JAAD and PIE as in~\cite{rasouli2019pie}. A 0.5-second (15 frame) observation length and 1.5-second (45 frame) prediction horizon were used for evaluation. For ETH-UCY, a standard leave-one-out approach based on scene was used per~\cite{gupta2018social,salzmann2020trajectron++}. We observed trajectories for 3.2 seconds (8 frames) and predicted the paths for the next 4.8 seconds (12 frames). We used hidden size 256 for all encoders and decoders in BiTraP across all datasets. All models were trained with batch size 128, learning rate  0.001, and an exponential LR scheduler~\cite{salzmann2020trajectron++} on a single NVIDIA TITAN XP GPU.

\subsection{Experiments on JAAD and PIE Datasets}\label{sec:jaad-pie} 

\begin{table*}[h!]
    \centering
    \scriptsize
    \caption{Results on JAAD and PIE datasets. The center row shows deterministic baselines including our ablation model BiTraP-D;  the bottom row shows our proposed multi-modal methods. NLL is not available for deterministic methods since they predict single trajectories. Lower values are better. }
    \label{tab:jaad_pie_results}
    \begin{tabular}{l|cccc|cccc}
        \toprule
        \multirow{3}{*}{Methods} & \multicolumn{4}{c|}{JAAD} & \multicolumn{4}{c}{PIE} \\
        \cmidrule{2-9}
        & $ADE$ & $C_{ADE}$ &  $C_{FDE}$  & $NLL$ & $ADE$ & $C_{ADE}$ & $C_{FDE}$ & $NLL$\\ 
        & (0.5/1.0/1.5s) & (1.5s) & (1.5s) & & (0.5/1.0/1.5s) & (1.5s) & (1.5s) & \\
        \midrule
        Linear~\cite{rasouli2019pie}  & 233/857/2303 & 1565 & 6111 & - & 123/477/1365 & 950 & 3983 & -\\
        LSTM~\cite{rasouli2019pie} & 289/569/1558 & 1473 & 5766 & - & 172/330/911 & 837 & 3352 & -\\
        B-LSTM~\cite{bhattacharyya2018long} & 159/539/1535 & 1447 & 5615 & - & 101/296/855 & 811 & 3259 & - \\ %
        FOL-X ~\cite{yao2019egocentric} & 147/484/1374 & 1290 & 4924 & - & 47/183/584 & 546 & 2303 & -\\
        PIE$_{traj}$~\cite{rasouli2019pie} & 110/399/1280 & 1183 & 4780 & - & 58/200/636 & 596 & 2477 & -\\
        PIE$_{full}$~\cite{rasouli2019pie} & - & - & - & - & -/-/556 & 520 & 2162 & -\\
        \midrule
        BiTraP-D & \textbf{93/378/1206} & \textbf{1105} & \textbf{4565} & - & \textbf{41/161/511} & \textbf{481} & \textbf{1949} & - \\
        BiTraP-NP (20) & \textbf{38/94/222} & \textbf{177} & \textbf{565} & 18.9 & \textbf{23/48/102} & \textbf{81} & \textbf{261} & 16.5 \\
        BiTraP-GMM (20) & 153/250/585 & 501 & 998 & \textbf{16.0} & 38/90/209 & 171 & 368 & \textbf{13.8}\\
        \bottomrule
    \end{tabular}
\end{table*}

\begin{figure*}[h!]
    \centering
    \begin{subfigure}[h]{0.24\textwidth}
        \includegraphics[width=0.95\textwidth]{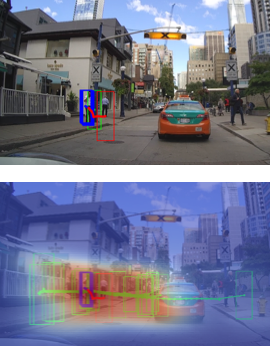}
        \caption{}
        \label{fig:result_PIE_1}
    \end{subfigure}
    \begin{subfigure}[h]{0.24\textwidth}
        \includegraphics[width=0.95\textwidth]{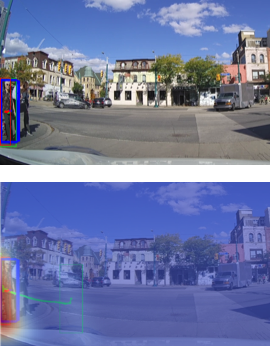}
        \caption{}
        \label{fig:result_PIE_2}
    \end{subfigure}
    \begin{subfigure}[h]{0.24\textwidth}
        \includegraphics[width=0.95\textwidth]{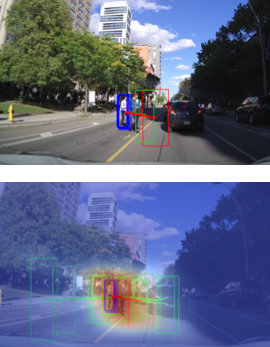}
        \caption{}
        \label{fig:result_PIE_3}
    \end{subfigure}
    \begin{subfigure}[h]{0.24\textwidth}
        \includegraphics[width=0.95\textwidth]{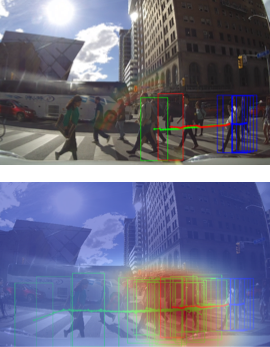}
        \caption{}
        \label{fig:result_PIE_4}
    \end{subfigure}
    \caption{Qualitative results of deterministic (top row) vs multi-modal (bottom row) bi-directional prediction. Past (dark blue), ground truth future (red) and predicted future (green) trajectories and final bounding box locations are plotted. In the bottom row, each BiTraP-NP likelihood heatmap fits a KDE over samples. The orange color indicates higher probability.}
    \label{fig:bitrap_np_PIE}
\end{figure*}

\xhdr{Baselines.} We compare our results against the following baseline models: 1) Linear Kalman filter, 2) Vanilla LSTM model, 3)  Bayesian-LSTM model (B-LSTM)~\cite{bhattacharyya2018long}, 4) PIE$_{traj}$, an attentive RNN encoder-decoder model, 5) PIE$_{full}$, a multi-stream attentive RNN model, by injecting ego-motion and semantic intention stream to PIE$_{traj}$, and 6) FOL-X~\cite{yao2019egocentric}, a multi-stream RNN encoder-decoder model using residual prediction. We also conducted an ablation study for a deterministic variation of our model (BiTraP-D), where the multi-modal CVAE module was removed.

\xhdr{Evaluation Metrics.} Following \cite{yao2019egocentric,rasouli2019pie,bhattacharyya2018long}, our BiTraP model was evaluated using: 1) bounding box Average Displacement Error ($ADE$), 2) box center ADE ($C_{ADE}$) and 3) box center Final Displacement 
Error ($C_{FDE}$) in squared pixels.
For our multi-modal BiTraP-NP and BiTraP-GMM, we compute the 
best-of-20 results (the minimum ADE and FDE from 20 randomly-sampled trajectories),
following~\cite{gupta2018social,salzmann2020trajectron++,sadeghian2019sophie}. 
We also report the Kernel Density Estimation-based Negative Log Likelihood (KDE-NLL) metric for BiTraP-NP and BiTraP-GMM, which evaluates the NLL of the ground truth under a distribution fitted by a KDE on 2000 trajectory 
samples from each prediction model~\cite{salzmann2020trajectron++, thiede2019analyzing}. For all metrics, lower values are better.

\xhdr{Results.} Table~\ref{tab:jaad_pie_results} presents trajectory prediction results with JAAD and PIE datasets. Our deterministic BiTraP-D model 
shows consistently lower displacement errors across various prediction horizons than baseline methods such as PIE$_{traj}$ and FOL-X 
indicating our goal estimation and bi-directional 
prediction modules are effective. Our BiTraP-D model, based only on past trajectory information, also outperforms the state-of-the-art PIE$_{full}$, which requires additional ego-motion and semantic intention annotations. 
Table~\ref{tab:jaad_pie_results} also shows that non-parametric multi-modal method BiTraP-NP performs better on 
displacement metrics while parametric method BiTraP-GMM performs better on the 
$NLL$ metric. This difference illustrates the objectives of these methods: BiTraP-NP generates diverse trajectories, and one trajectory was optimized to have minimum displacement error, while BiTraP-GMM generates trajectory distributions with more similarity to the ground truth trajectory.

\begin{table*}[ht!]
    \centering
    \caption{Trajectory prediction results (ADE/FDE) on BEV ETH-UCY datasets. Lower is better.}
    \begin{tabular}{c|ccccc|cc}
        \toprule
        Datasets & S-GAN~\cite{gupta2018social} & SoPhie~\cite{sadeghian2019sophie} & S-BiGAT~\cite{kosaraju2019social} & PECNet~\cite{mangalam2020endpoint} &Trajectron++~\cite{salzmann2020trajectron++}  & BiTraP-NP & BiTraP-GMM \\ 
        \midrule
        ETH & 0.81/1.52 & 0.70/1.43 & 0.69/1.29 & 0.54/0.87 & 0.43/0.86  & \textbf{0.37/0.69}  & 0.40/0.74\\ 
        Hotel & 0.72/1.61 & 0.76/1.67 & 0.49/1.01 & 0.18/0.24 & \textbf{0.12/0.19} & \textbf{0.12}/0.21  & 0.13/0.22\\
        Univ & 0.60/1.26 & 0.54/1.24 & 0.55/1.32 & 0.35/0.60 & 0.22/0.43 & \textbf{0.17/0.37} & 0.19/0.40\\ 
        Zara1 & 0.34/0.69 & 0.30/0.63 & 0.30/0.62 & 0.22/0.39 & 0.17/0.32 & \textbf{0.13/0.29}  & 0.14/\textbf{0.28}\\  
        Zara2 & 0.42/0.84 & 0.38/0.78 & 0.36/0.75 & 0.17/0.30 & 0.12/0.25 & \textbf{0.10/0.21} & 0.11/0.22\\
        \midrule 
        Average & 0.58/1.18 & 0.54/1.15 & 0.48/1.00 & 0.29/0.48 & 0.21/0.39 & \textbf{0.18/0.35} & 0.19/0.37\\
        \bottomrule
    \end{tabular}
    \label{tab:eth_ucy_bom20}
\end{table*}

\begin{figure*}[htbp]
    \centering
    \begin{subfigure}[h]{0.24\textwidth}
        \includegraphics[width=0.85\textwidth]{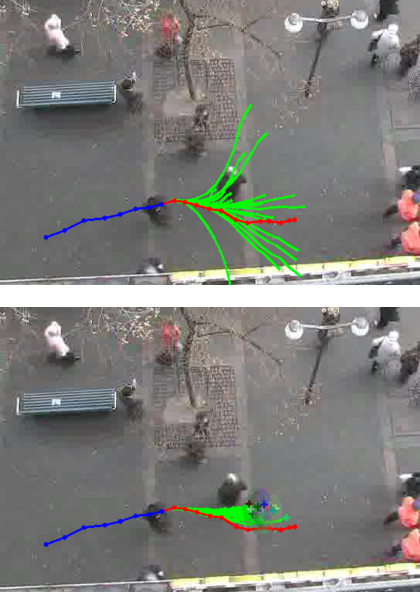}
        \caption{Hotel}
        \label{fig:result_hotel}
    \end{subfigure}
    \begin{subfigure}[h]{0.24\textwidth}
        \includegraphics[width=0.85\textwidth]{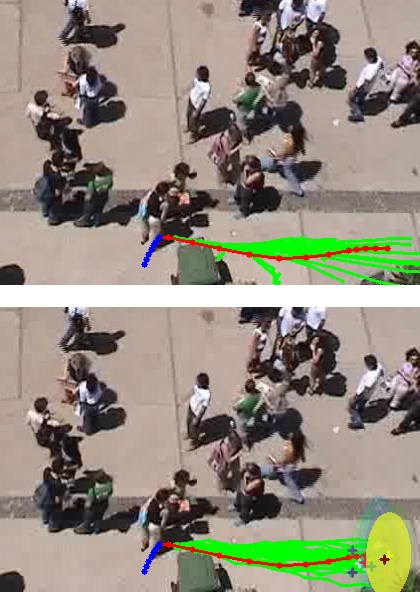}
        \caption{Univ}
        \label{fig:result_univ}
    \end{subfigure}
    \begin{subfigure}[h]{0.24\textwidth}
        \includegraphics[width=0.85\textwidth]{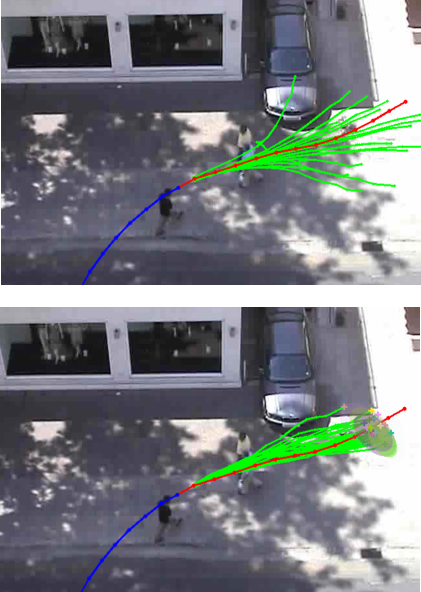}
        \caption{Zara2}
        \label{fig:result_zara2}
    \end{subfigure}
    \begin{subfigure}[h]{0.24\textwidth}
        \includegraphics[width=0.85\textwidth]{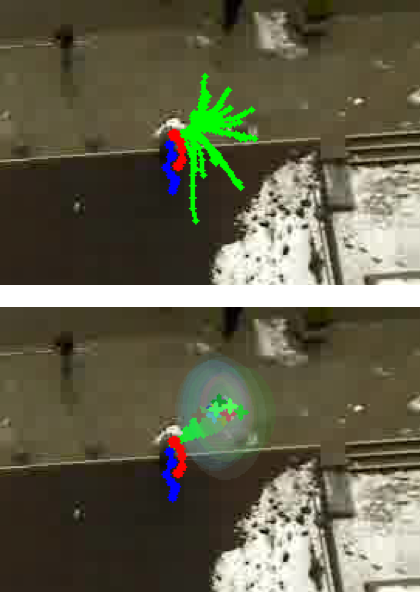}
        \caption{ETH}
        \label{fig:result_eth}
    \end{subfigure}
    \caption{Visualizations of BiTraP-NP (first row) and BiTraP-GMM (second row). Twenty sampled future trajectories are plotted. For BiTraP-GMM, we also plot end-point GMM distributions as colored ellipses. Size indicates component $\Sigma_k$ and transparency indicates component weight $\pi_k$. }
    \label{fig:qualitative_eth}
\end{figure*}

\begin{table}[h!]
    \centering
    \footnotesize
    \caption{Average-NLL/Final-NLL (ANLL/FNLL) results on ETH-UCY datasets. Lower is better.}
    \resizebox{.49 \textwidth}{!} {
    \begin{tabular}{c|cc|cc}
        \toprule
        Datasets & S-GAN~\cite{gupta2018social} &Trajectron++~\cite{ivanovic2019trajectron} & BiTraP-NP & BiTraP-GMM  \\
        \midrule
        ETH & 15.70/- & 1.31/4.28 & 3.80/3.79 & \textbf{0.96/3.55}\\
        Hotel & 8.10/- & \textbf{-1.94/0.25} &-0.41/1.26 & -1.60/0.51\\
        Univ & 2.88/- & -1.13/2.13  & -0.84/2.15 & \textbf{-1.19/2.03} \\
        Zara1 & 1.36/- & -1.41/1.83 & -0.81/1.85 & \textbf{-1.51/1.56}\\
        Zara2 & 0.96/- & -2.53/0.50  & -1.89/1.31 & \textbf{-2.54/0.38}\\%
        \bottomrule
    \end{tabular}}
    \label{tab:nll}
    \vspace{-12pt}
\end{table}
Fig.~\ref{fig:bitrap_np_PIE} shows trajectory prediction results on sample frames from the PIE dataset. We observed that when a pedestrian intends to cross the street or change directions, the multi-modal BiTraP methods yield higher accuracy and more reasonable predictions than the deterministic variation. For example, as shown in Fig.~\ref{fig:result_PIE_2}, the deterministic BiTraP-D model (top row) can fail to predict the trajectory and the end-goal, where a pedestrian intends to cross the street in the future; the multi-modal BiTraP-NP model (bottom row) can successfully predict multiple possible future trajectories, including one where the pedestrian is crossing the street matching ground truth intention. Similar observations can be made in other frames. This result indicates  multi-modal BiTraP-NP can predict multiple possible futures, which could help a mobile robot or a self-driving car safely yield to pedestrians.
Although BiTraP-NP samples diverse trajectories, it still predicts distribution with high likelihood around ground truth targets and low likelihood in other locations as shown in Fig.~\ref{fig:result_PIE_2}-\ref{fig:result_PIE_4}.

\subsection{Experiments on ETH-UCY Datasets}\label{sec:eth-ucy}

\xhdr{Baselines.} We compare our methods with five multi-modal baseline methods: S-GAN~\cite{gupta2018social}, SoPhie~\cite{sadeghian2019sophie}, S-BiGAT~\cite{kosaraju2019social}, PECNet~\cite{mangalam2020endpoint} and Trajectron++~\cite{salzmann2020trajectron++}. PECNet and Trajectron++ are most recent. PCENet is a goal-conditioned method using non-parametric distribution (thus directly comparable to our BiTraP-NP) while Trajectron++ uses a GMM trajectory distribution directly comparable to our BiTraP-GMM. Note that the baselines incorporated social information while our method focuses on investigating goal-based trajectory modeling and do not require extra input such as social or scene information. 

\xhdr{Evaluation Metrics.} Following~\cite{gupta2018social,mangalam2020endpoint,sadeghian2019sophie}, we used best-of-20 trajectory ADE and FDE in meters as evaluation metrics. We also report Average and Final KDE-NLL (ANLL and FNLL) metrics on 2000 sampled trajectories~\cite{thiede2019analyzing,salzmann2020trajectron++} to evaluate the predicted trajectory and goal distribution.

\xhdr{Results.} Table~\ref{tab:eth_ucy_bom20} shows the best-of-20 ADE/FDE results across all methods. We observed that BiTraP-NP outperforms the state-of-the-art goal based method (PECNet) by a large margin ($\sim12\%-51\%$), demonstrating the effectiveness of our bi-directional decoder module. BiTraP-NP also obtains lower ADE/FDE on most scenes ($\sim12\%$-$24\%$ improvement) compared with Trajectron++. Our BiTraP-GMM model was trained using NLL loss, so it shows higher ADE/FDE results compared with BiTraP-NP. This is consistent with our FPV dataset observations in Section~\ref{sec:jaad-pie}. Nevertheless, BiTraP-GMM still achieves similar or better results than PECNet and Trajectron++. 

\begin{table*}[htbp]
    \centering
    \caption{Ablation study results (ADE/FDE and ANLL/FNLL). Lower is better.}
    \resizebox{.9 \textwidth}{!} {
    \begin{tabular}{c|cccc|cccc}
        \toprule
        \multirow{3}{*}{Method}& \multicolumn{4}{c|}{BiTraP-NP} & \multicolumn{4}{c}{BiTraP-GMM}\\
        & \multicolumn{2}{c}{w/o backward (TraP-NP)} & \multicolumn{2}{c|}{w/ backward} &  \multicolumn{2}{c}{ w/o bi-loss} & \multicolumn{2}{c}{w/ bi-loss}\\
        \midrule
         & ADE/FDE & ANLL/FNLL & ADE/FDE & ANLL/FNLL & ADE/FDE & ANLL/FNLL & ADE/FDE & ANLL/FNLL\\
        ETH & 0.44/0.96 & 4.20/4.45  & \textbf{0.37/0.69} & \textbf{3.80/3.79}
 & 0.43/0.80 & 1.11/3.81 & \textbf{0.40/0.74} & \textbf{0.96/3.55}\\ 
        Hotel & 0.13/0.23 & -0.17/1.64 & \textbf{0.12/0.21} & \textbf{-0.41/1.26} & 0.16/0.25 & -1.32/0.80 & \textbf{0.13/0.22} & \textbf{-1.60/0.51}\\
        Univ & 0.21/0.43 &  -0.21/2.78& \textbf{0.17/0.37} & \textbf{-0.84/2.15} & 0.20/0.41 & -1.16/2.06 & \textbf{0.19/0.40} & \textbf{-1.19/2.03}\\ 
        Zara1 & 0.15/0.31 & -0.37/2.27 & \textbf{0.13/0.29} & \textbf{-0.81/1.85}
& 0.19/0.35 & -0.90/2.12 & \textbf{0.14/0.28} & \textbf{-1.51/1.56}\\  
        Zara2 & 0.12/0.23 & -1.70/1.54& \textbf{0.10/0.21} & \textbf{-1.89/1.31} & 0.13/0.25 &  -2.38/0.64 & \textbf{0.11/0.22} & \textbf{-2.54/0.38}\\
        \bottomrule
    \end{tabular}}
    \label{tab:ablation}
    \vspace{-5pt}
\end{table*}

\begin{table}[htbp]
    \centering
    \caption{Computational times with 20/2000 samples.}
    \resizebox{.49 \textwidth}{!} {
    \begin{tabular}{c|ccc}
        \toprule
        Method & Scene Graph & Model inference & Total\\
        \midrule
        S-GAN\cite{gupta2018social} & N/A & 103/10445 ms & 103/10300 ms\\
        Trajectron++\cite{salzmann2020trajectron++} & 11ms & 55/58 ms & 66/69 ms\\
        \midrule
        TraP-NP & N/A & 5.3/5.9 ms & 5.3/5.9 ms\\
        BiTraP-NP & N/A  & 8.3/9.1 ms & 8.3/9.1 ms\\
        BiTraP-GMM & N/A & 69/72ms & 69/72ms\\
        \bottomrule
    \end{tabular}}
    \label{tab:time}
    \vspace{-8pt}
\end{table}

To further evaluate predicted trajectory distributions, 
we report KDE-NLL results in Table~\ref{tab:nll}. As shown, BiTraP-GMM outperforms Trajectron++ with lower ANLL and FNLL on \textit{ETH}, \textit{Univ}, \textit{Zara1} and \textit{Zara2} datasets. On \textit{Hotel}, Trajectron++ achieves lower NLL values which may be due to the possible higher levels of inter-personal interactions than in other scenes. 
We observed improved ANLL/FNLL on \textit{Hotel} (-1.88/0.27) when combining the BiTraP-GMM decoder with the interaction encoder in~\cite{salzmann2020trajectron++}, consistent with our hypothesis.

Fig.~\ref{fig:qualitative_eth} shows qualitative examples of our predicted trajectories using the BiTraP-NP and BiTraP-GMM
models. As shown, BiTraP-NP (top row) generates future possible trajectories with a wider spread (more diverse), while BiTraP-GMM generates more compact distributions. This is consistent with our quantitative evaluations as reported in Table~\ref{tab:nll}, where the lower NLL results of BiTraP-GMM correspond to more compact trajectory distributions.

\begin{figure}[htbp]
    \centering
    \begin{subfigure}[h]{0.49\textwidth}
        \centering
        \includegraphics[width=0.85\textwidth]{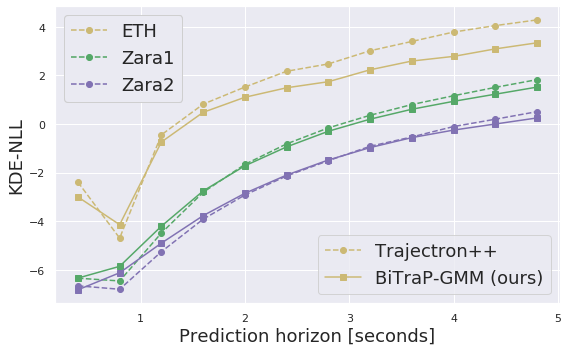}
        \label{fig:per_step_kde_nll_1}
    \end{subfigure}
    \begin{subfigure}[h]{0.49\textwidth}
        \centering
        \includegraphics[width=0.85\textwidth]{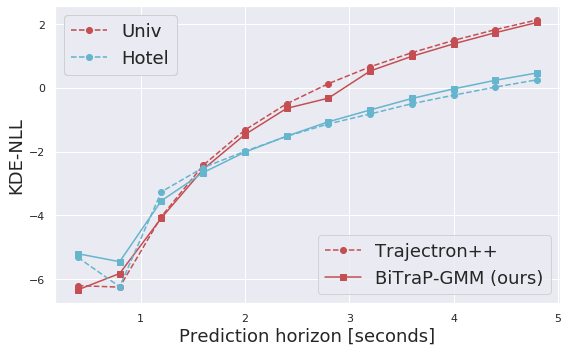}
        \label{fig:per_step_kde_nll_2}
    \end{subfigure}
    \caption{KDE-NLL results on the ETH-UCY dataset per timestep up to 4.8 seconds.} 
    \label{fig:per_step_results}
    \vspace{-15pt} 
\end{figure}


We also computed KDE-NLL results for both Trajectron++ and BiTraP-GMM methods at each time step to analyze how BiTraP affects both short-term and longer-term (up to 4.8 seconds) prediction results. Per Fig.~\ref{fig:per_step_results}, BiTraP-GMM outperforms Trajectron++ with longer prediction horizons (after 1.2 seconds on \textit{ETH}, \textit{Univ}, \textit{Zara1}, and \textit{Zara2}). This shows the backward passing from the goal helps reduce error with longer prediction horizon. 

\subsection{Additional Experiments}
\xhdr{Ablation study.} We conducted two ablation experiments. To show bi-directional decoder effectiveness, we removed the backward decoder from BiTraP-NP and compared its performance with the original BiTraP-NP model (w/o backward (TraP-NP) vs w/ backward). To show bi-directional loss effectiveness in BiTraP-GMM, we compared two BiTraP-GMM models trained with forward loss and bi-directional loss (w/o bi-loss vs w/ bi-loss).  A comparison of ADE/FDE and ANLL/FNLL results is presented in Table~\ref{tab:ablation}. Using a bi-directional decoder (BiTraP-NP) improves ADE/FDE by 10\%-28\% (ANLL/FNLL by $\sim$0.4) from the model without backward decoder. By using bi-directional loss (bi-loss), the ADE/FDE of BiTraP-GMM model improves by 5-18\% on ETH, and ANLL/FNLL improves by $\sim$0.25.

\xhdr{Computational time.} We provide model inference time of Social GAN~\cite{gupta2018social}, Trajectron++~\cite{salzmann2020trajectron++} and our BiTraP-NP and BiTraP-GMM models in Table~\ref{tab:time}. Trajectron++ generates scene graphs before running the model so computation time is summed over scene graph generation and model inference. For Social GAN and our method, total time consists of model inference time only. We show computational times for number of samples 20 and 2000. Time differences of BiTraP models between the two numbers are $\sim3$ms, while the difference of S-GAN is extremely large as it generates samples one-by-one. BiTraP-GMM is $\sim3$ms slower than Trajectron++, not significant since both methods run at $\sim70$ms per frame ($\sim14$ FPS) on average. BiTraP-NP is about 8x faster than Trajectron++ and BiTraP-GMM since it does not fit a GMM model or perform dynamic integration. Adding the bi-directional decoder slows inference by $\sim3$ms (TraP-NP vs BiTraP-NP). All experiments are conducted on the same machine used for training.

\xhdr{Robot Navigation Simulation Experiment.} To intuitively present model performance in robot navigation, we conducted a robot navigation simulation experiment on the ETH-UCY dataset. We observed that a robot equipped with the BiTraP-NP predictor is more sensitive to potential collisions as it predicts diverse distributions of surrounding pedestrians. On the other hand, a robot equipped with BiTraP-GMM predictor reports less false alarm when navigating among pedestrians and, thus, being more efficient. Experimental details can be found in the supplementary material.

\section{Conclusion}
We presented \textit{BiTraP}, a bi-directional multi-modal trajectory prediction method conditioned on goal estimation. We demonstrated that our proposed model can achieve state-of-the-art results for pedestrian trajectory prediction on both first-person view and bird's eye view datasets. The current BiTraP models, with only observed trajectories as inputs, already surpass previous methods which required additional ego-motion, semantic intention, and/or social information. By conducting a comparative study between non-parametric (BiTraP-NP) and parametric (BiTraP-GMM) models, we observed that the different latent variable choice affects the diversity of target distributions of future trajectories. 
We hypothesized that such difference in predicted distribution directly influences the collision rate in robot path planning and showed that collision metrics can be used to guide predictor selection in real world applications.
For future work, we plan to incorporate scene semantics and social components to further boost the performance of each module. We are also interested in using predicted goals and trajectories to infer and interpret pedestrian intention and actions.






\bibliographystyle{IEEEtran} 
\bibliography{Reference_ieee} 

\end{document}


\maketitle
\section{CVAE Preliminaries}
A Conditional Variation Autoencoder (CVAE) is a conditional generative model designed to output target data $Y$ based on latent variable $Z$ and observation $X$~\cite{sohn2015learning}. A CVAE consists of three modules: a \textbf{conditional prior network} $p_\theta(Z|X)$ to model latent variable $Z$ conditioned on observation $X$, a \textbf{recognition network} $q_\phi(Z|X, Y)$ to capture dependencies between $Z$ and target $Y$, and a \textbf{generation network} $p_\psi(Y|X, Z)$ to generate the target $Y$, where $\phi$, $\theta$, and $\psi$ represent network parameters. Stochastic latent variable $Z\in\mathbb{R}^d$ is sampled from a pre-defined distribution format such as a Gaussian distribution. The CVAE samples $Z$ and generates target $Y$ conditioned on observation $X$. The objective of a typical CVAE model is to maximize its variational lower bound
\begin{align}
\begin{split}
\label{eq:vlb}
    \max_{\theta,\phi,\psi}  &\mathbb{E}_{
    q_\phi(Z|X,Y)}
    \Big[
        \log{p_\psi(Y|X,Z)}
    \Big]\\
    &-KL\Big(q_\phi(Z|X,Y)||p_\theta(Z|X)\Big),
    \end{split}
\end{align}
where the first term maximizes the expectation of the log-likelihood 
of the target in the predicted distribution; 
the $K$-$L$ (Kullback–Leibler) divergence term minimizes the difference 
between the recognition network and the conditional prior network. In this paper, we designed a modified CVAE with two generation networks and optimize both networks end-to-end.

\section{Robot Navigation Simulation Experiment Using BiTraP}

To quantitatively analyze application of the BiTraP-GMM and BiTraP-NP models to robot navigation tasks, we designed a simulated robot navigation
experiment based on the ETH-UCY bird's-eye view dataset. In this experiment, given  predicted pedestrian trajectory distributions in a scene using our BiTraP models and pre-planned paths for a robot, we show that we are able to compute the collision likelihood for each path, and thus are able to predict collision rate and select the safest path for the robot. 
Assuming a mobile robot navigates among pedestrians, we present results on two tasks: 1) Select the safest path for the robot and 2) Predict whether a path will collide with any other pedestrians in the scene. In this section, we first introduce our experiment setup. Then, we present evaluation results of our BiTraP models on path selection and collision prediction tasks.

\begin{figure}[htbp]
    \centering
    \includegraphics[width=0.48\textwidth]{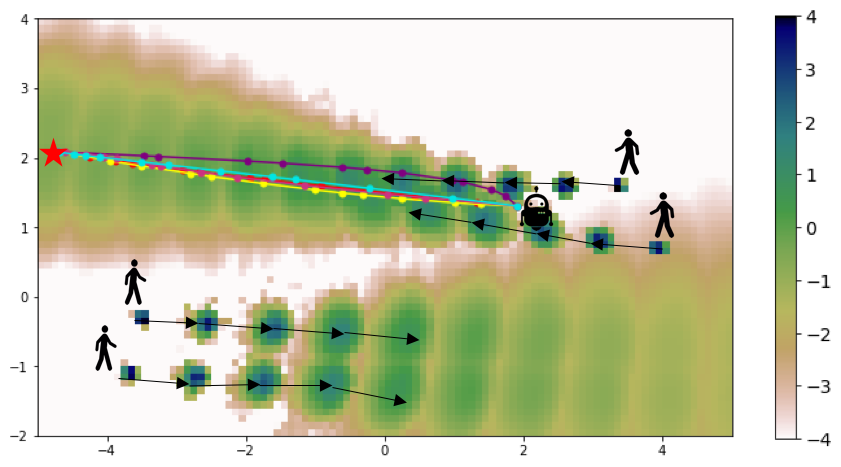}
    \caption{Generation of Monte Carlo (MC) robot trajectories for collision detection experiments using Bezier curves. We illustrate five MC trajectory samples including start (robot icon) and end (red star) waypoints. Predicted trajectory distributions of neighbor pedestrians are plotted as a heat map; their walking directions are indicated by black arrows.}
    \label{fig:simulation_exp}
\end{figure}

\xhdr{Experimental Setup.} We selected all samples with more than one pedestrian in the test split~\cite{gupta2018social} from ETH-UCY. Each sample has a node pedestrian (the pedestrian used for testing in previous work) and several neighbor pedestrians (the pedestrians used for social modeling in previous work) as
in~\cite{gupta2018social,salzmann2020trajectron++}. 
We regard the node pedestrian as a "robot" navigating among other neighbor pedestrians. The starting and goal points of the "robot" are the same as the current position and goal point of the node pedestrian. A sample scene with one ``robot'' navigating among four other pedestrians is shown in Fig.~\ref{fig:simulation_exp}. For the robot, $100$ Monte Carlo (MC) paths were generated from start state to end point following quadratic and cubic Bezier curves~\cite{gallier2000curves}. Other more complex path planners could be used to generate additional experimental datasets. We assume the robot must reach the designated goal in 12 time steps, matching the prediction horizon for the pedestrian node in each scene. We uniformly generate waypoints along the path and randomly shift each by up to $\pm50\%$ of the step length, resulting in a trajectory sequence containing 12 random waypoints. Other pedestrians follow their original (ground truth) trajectories in the scene. For each neighbor pedestrian, we run BiTraP-NP and BiTraP-GMM separately. Each method samples 2000 future trajectories to fit one Gaussian Kernel Density Estimation (KDE) model for each pedestrian as the predicted future distribution. 
Then, we compute the maximum KDE log-likelihood of all the waypoints on all robot MC paths and treat this log-likelihood value as a collision score. The higher the collision score, the more likely a collision will happen along this path. Given these collision scores, we compute the safest path collision rate (SPCR) as reported in \textit{Task 1} below.  Receiver operating characteristic (ROC) and precision-recall (P-R) curve results are reported in \textit{Task 2}. 

\xhdr{Task 1: Predict the Safest Path.} We mark the robot MC path in each scene with minimum collision score as the ``safest" (lowest collision likelihood) path. Then, we compute Euclidean distances between each safest path waypoint and other pedestrians' ground truth future trajectories. A collision is tallied if the minimum distance between a path and any pedestrians in the scene is less than $0.2$ meters. Collision rate is computed as the number of paths with collision divided by the total number of safest paths. Due to the randomness in MC path generation, we conducted the simulation experiment five times with BiTraP-NP and BiTraP-GMM predictors separately and report collision rate mean ($\mu$) and standard deviation ($\sigma$) values in Table~\ref{tab:spcr_auc_ap}. As a comparison, we also present the collision rate of a randomly selected path among the 100 MC paths. The randomly selected paths do not have very high collision rates since the paths are planned based on pedestrian ground truth start and goal positions which are less likely to be involved in a collision. Compare to randomly selected paths, paths selected by our methods reduce the SPCR by a large margin. This shows that our predictors are effective for safest path selection. Both of our BiTraP methods achieve collision rate lower than $1\%$ on \textit{ETH}, \textit{Hotel} and \textit{Zara1} datasets. The \textit{Univ} dataset is more difficult due to its high pedestrian density, and \textit{Zara2} is most difficult because many pedestrian trajectories are quite close to each other. 
BiTraP-GMM shows lower SPCRs than BiTraP-NP on four datasets, indicating that it predicted more accurate (compared to ground truth) distributions. On \textit{Zara1}, BiTraP-NP outperforms BiTraP-GMM by a small margin. BiTraP-NP ANLL and FNLL metric values as reported in the main paper are still higher than BiTraP-GMM values. A possible explanation is that BiTraP-NP predicts more diverse distributions thus detects some collisions not identified by BiTraP-GMM.

\begin{table}[h]
    \centering
        \caption{SPCR($\mu\pm\sigma$), AUC and AP results of our methods on ETH-UCY data group.}
        \resizebox{.49 \textwidth}{!} {
    \begin{tabular}{c|c|c|c}
        \toprule
         & Random from 100 & BiTraP-NP & BiTraP-GMM \\
         & (SPCR) & (SPCR/AUC/AP) & (SPCR/AUC/AP) \\
        \midrule
        ETH & $0.6\pm0.4\%$ & $0.3\pm0.1\%/\,92.3/\,24.2$& $\mathbf{0.1\pm0.1\%}/\,\mathbf{95.5}/\,\mathbf{26.0}$ \\
        HOTEL & $0.4\pm0.3\%$ & $0.1\pm0.1\%/\,86.4/\,22.4$ & $\mathbf{0.0\pm0.0\%}/\,\mathbf{91.6}/\,\mathbf{29.1}$ \\
        Univ & $8.5\pm1.4\%$ & $5.8\pm0.5\%/\,81.0/\,33.4$ & $\mathbf{3.6\pm0.2\%}/\,\mathbf{87.6}/\,\mathbf{43.4}$ \\
        Zara1 & $2.4\pm0.5\%$ & $\mathbf{0.6\pm0.2\%}/\,88.9/\,38.6$ & $0.8\pm0.3\%/\,\mathbf{90.4}/\,\mathbf{41.6}$ \\
        Zara2 & $6.1\pm0.6\%$ & $3.2\pm0.1\%/\,81.0/\,44.0$ & $\mathbf{2.5\pm0.3\%}/\,\mathbf{87.5}/\,\mathbf{52.6}$ \\
        \bottomrule
    \end{tabular}}
    \label{tab:spcr_auc_ap}
\end{table}

\begin{figure}[h!]
    \begin{subfigure}[h]{0.48\textwidth}
        \includegraphics[width=0.95\textwidth]{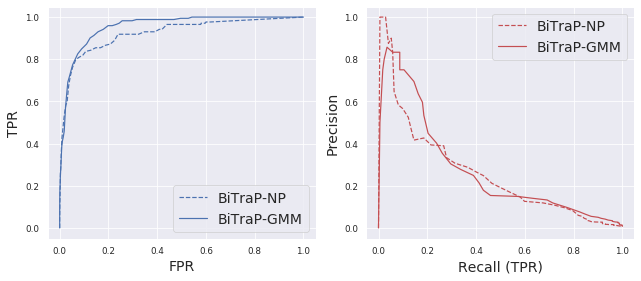}
        \caption{ETH}
        \label{fig:eth_roc_pr}
    \end{subfigure}
    \begin{subfigure}[h]{0.48\textwidth}
        \includegraphics[width=0.95\textwidth]{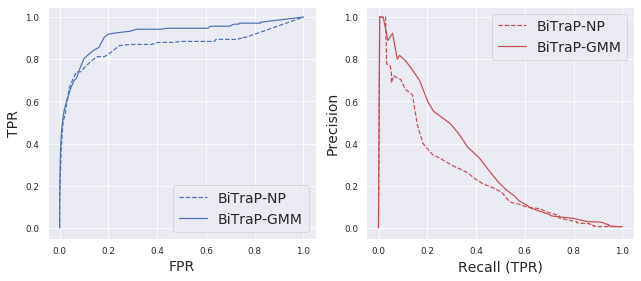}
        \caption{Hotel}
        \label{fig:hotel_roc_pr}
    \end{subfigure}
    
    \begin{subfigure}[h]{0.48\textwidth}
        \includegraphics[width=0.95\textwidth]{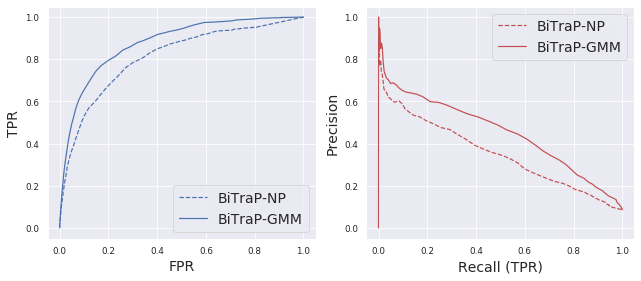}
        \caption{Univ}
        \label{fig:univ_roc_pr}
    \end{subfigure}
    \begin{subfigure}[h]{0.48\textwidth}
        \includegraphics[width=0.95\textwidth]{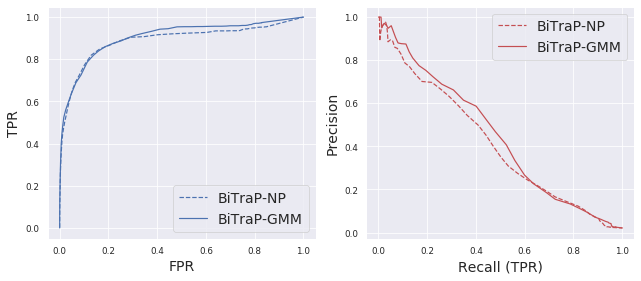}
        \caption{Zara1}
        \label{fig:zara1_roc_pr}
    \end{subfigure}
    
    \begin{subfigure}[h]{0.48\textwidth}
        \includegraphics[width=0.95\textwidth]{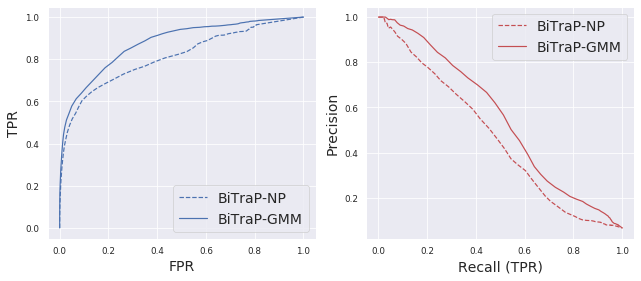}
        \caption{Zara2}
        \label{fig:zara2_roc_pr}
    \end{subfigure}
    \caption{ROC (left) and P-R (right) curves of BiTraP-NP and BiTraP-GMM on ETH dataset.}
    \label{fig:roc_pr}
\end{figure}

\begin{table*}[ht!]
    \centering
    \begin{tabular}{l|cccc|cccc}
        \toprule
        \multirow{2}{*}{Method} & \multicolumn{4}{c|}{KDE NLL} & \multicolumn{4}{c}{FDE ML} \\
        \cmidrule{2-9} 
        & @1s & @2s & @3s & @4s & @1s & @2s & @3s & @4s \\
        \midrule
        Trajectron++ base~\cite{salzmann2020trajectron++} & -2.69 & -2.46 & -1.76 & -1.09 & 0.03 & 0.17 & 0.37 & 0.60 \\
        Trajectron++ $\int$, map~\cite{salzmann2020trajectron++} & -5.58 & -3.96 & -2.77 & -1.89 & \textbf{0.01} & 0.17 & 0.37 & 0.62 \\
        BiTraP-GMM (ours) & \textbf{-6.08} & \textbf{-4.21} & \textbf{-2.98} & \textbf{-2.05} & 0.02 & \textbf{0.15} & \textbf{0.35} & \textbf{0.58} \\
        \bottomrule
    \end{tabular}
    \caption{Pedestrian-only trajectory prediction results on nuScenes dataset.}
    \label{tab:nuscenes_ped}
\end{table*}
\xhdr{Task 2: Predict Collision for Any Path.} The collision rate metric above only evaluates the safest path as selected by a trajectory predictor thus neglects all other paths. In the real-world, a trajectory predictor must be sufficiently accurate for the robot to accurately predict future collisions with high precision with a low missing rate (high true positive rate, TPR) and a low false alarm rate (low false positive rate, FPR). To show the performance of BiTraP-NP and BiTraP-GMM predictors in terms of these metrics, we plotted the collision prediction ROC curve and P-R curve as follows. First, we collected all MC paths for the robot and tallied their collision scores. By setting a threshold $\gamma$, we can classify a path as collided (positive) or not collided (negative) and compute the TPR (i.e., recall), FPR and precision values. The ground truth label of each path is computed in the same way as before. By decreasing $\gamma$ from a maximum value to minimum value (6 and -10 in this work), we plot the ROC and P-R curves shown in Fig.~\ref{fig:roc_pr}. The corresponding area under curve (AUC) and average precision (AP) are presented in Table~\ref{tab:spcr_auc_ap}. In this work, AP is computed by equally spaced recall levels \{1/40, 2/40,...,1\} following~\cite{simonelli2019monodis}.

As shown in Fig.~\ref{fig:roc_pr} and Table~\ref{tab:spcr_auc_ap}, both BiTraP-NP and BiTraP-GMM methods achieve high AUCs (e.g., $>90$ on \textit{ETH}). Generally, BiTraP-GMM outperforms BiTraP-NP by a small margin in terms of both AUC and AP (e.g., 95.5 vs 92.3 AUC, and 26.0 vs 24.2 AP on \textit{ETH}). Note that in real-world mobile robot applications missed collision detection (false negative) is unacceptable due to safety. That is to say, a high TPR (recall) is required. As can be observed in the higher TPR regions (x-axis) of the P-R curves, BiTraP-GMM outperforms BiTraP-NP on \textit{ETH} (Fig.~\ref{fig:eth_roc_pr}) and \textit{Hotel} (Fig.~\ref{fig:hotel_roc_pr}), and both methods perform similarly on \textit{Zara1} (Fig.~\ref{fig:zara1_roc_pr}). On \textit{Univ} (Fig.~\ref{fig:univ_roc_pr}) and \textit{Zara2} (Fig.~\ref{fig:zara2_roc_pr}), when the TPR is greater than a relatively high value (say 0.8), the FPR are higher  ($>0.2$) than in the other datasets, indicating increased chance of false alarms on these two datasets. 

Compared to the ROC curve, the P-R curve is more suitable for imbalanced datasets  due to the fact that it evaluates the fraction of true positives among positive predictions. This fits our case where the ratio of with-collision to no-collision paths is around 1:140, a large imbalance. On \textit{Univ} and \textit{Zara2} (Fig.~\ref{fig:univ_roc_pr} and~\ref{fig:zara2_roc_pr}), BiTraP-GMM has higher precision than BiTraP-NP across almost all recall values. On the other hand, on \textit{ETH}, \textit{Hotel} and \textit{Zara1} (Fig.~\ref{fig:eth_roc_pr}~\ref{fig:hotel_roc_pr} and~\ref{fig:zara1_roc_pr}), the two methods achieve similar precision at higher recall regions (e.g., when recall$>0.6$). This is because when the threshold $\gamma$ is too low, many paths are predicted as collided by both methods. 

The ROC and P-R curves also verified our observation regarding the diversity of the predicted trajectory distribution as described in the main paper. At a fixed TPR on the ROC curves, we observe that BiTraP-NP always has a greater FPR than BiTraP-GMM, consistent with our hypothesis that BiTraP-NP predicts more diverse distributions,  thus predicts more false alarms. Similarly, with fixed recall in P-R curves, BiTraP-NP has lower precision due the greater number of false alarms.

In summary, this simulated robot collision experiment demonstrated our proposed BiTraP trajectory predictor can be used in future robotic applications, such as predicting collisions and selecting safest paths in robot navigation tasks. Results from this supplementary experiment are consistent with our main paper's observations and further verify our hypothesis regarding the diversity/compactness of predicted trajectory distributions, i.e., BiTraP-NP predicts more diverse distributions while BiTraP-GMM predicts more compact distributions. The SPCR, ROC (AUC) and P-R (AP) metrics used in this experiment act as a supplement to the currently reported and widely used ADE/FDE and KDE-NLL metrics in the main paper. We believe these additional metrics and experiments offer an intuitive and complementary performance evaluation of the two proposed BiTraP models (NP and GMM) and their applications for tasks such as collision prediction and path selection. 

\section{Experiment and Result on nuScenes Dataset}
Among the datasets we have evaluated on, JAAD and PIE are first-person view only while ETH and UCY are focusing on campus or sidewalks only. To further present the performance of BiTraP in bird's eye view autonomous driving scenarios, we evaluate on the nuScenes dataset~\cite{nuscenes2019}. The nuScenes dataset contains trajectories collected from 850 scenes, 700 for training and 150 for testing~\cite{nuscenes2019}. We followed~\cite{salzmann2020trajectron++} to extract training and testing trajectories and trained our model using the same configurations as in ETH-UCY experiment. Note that we treat the pedestrian position at 4 seconds in the future as the target of our goal or end-point during training. 

\xhdr{Evaluation metrics.} To be comparable with~\cite{salzmann2020trajectron++}, the most-likely (ML) prediction is used to compute the final displacement error (FDE). We also use the kernel density estimation negative log-likelihood (KDE NLL) as in our other experiments.

\begin{table}[htbp]
    \centering
    \begin{tabular}{l|cccc}
        \toprule
        \multirow{2}{*}{Method} & \multicolumn{4}{c}{FDE ML} \\
        \cmidrule{2-5} 
         & @1s & @2s & @3s & @4s \\
        \midrule
        Trajectron++ base~\cite{salzmann2020trajectron++} & 0.18 & 0.57 & 2.25 & 2.24 \\
        Trajectron++ $\int$, map~\cite{salzmann2020trajectron++} & \textbf{0.07} & 0.45 & 1.14 & 2.20 \\
        BiTraP-GMM (ours) & 0.08 & \textbf{0.43} & \textbf{1.06} & \textbf{1.99} \\
        \bottomrule
    \end{tabular}
    \caption{Vehicle-only trajectory prediction results on nuScenes dataset.}
    \label{tab:nuscenes_veh}
\end{table}

\xhdr{Results.} As can be seen in Table~\ref{tab:nuscenes_ped}, adding dynamic integration and map encoding to the base Trajectron++ improved the distribution accuracy by a large margin but does not affect the FDE ML, indicating similar modes but smaller variances of the predicted distributions. Trajectron++ based methods used interactions and/or encoded map as inputs while our BiTraP-GMM only takes target pedestrians past trajectory. As in Table~\ref{tab:nuscenes_ped}, BiTraP-GMM improves the KDE-NLL at all evaluated time steps and also improves FDE after 2 seconds, showing how does the bi-directional strategy improves prediction accuracy. Note that the Trajectron++ benchmark lacks a ablation with integration but not map encoding (e.g. Trajectron++ $\int$) to show the necessity of map. However, our experiment shows that map may not be a very important information when predicting pedestrian trajectories on nuScenes dataset since BiTraP-GMM outperforms ``Trajectron++ $\int$, map".

\bibliographystyle{IEEEtran} 
\bibliography{Reference_ieee}